\documentclass[11pt]{article}
\usepackage{acl2016}
\usepackage{times}
\usepackage{url}
\usepackage{latexsym}
\usepackage{graphicx}
\usepackage{mathtools}
\usepackage{scrextend}
\usepackage{multirow}
\usepackage{array,booktabs}
\usepackage{enumitem}
\usepackage{csquotes}
\usepackage{dsfont}

\aclfinalcopy 

\title{Towards End-to-End Learning for Dialog State Tracking and Management using Deep Reinforcement Learning}

\author{Tiancheng Zhao and Maxine Eskenazi \\
  Language Technologies Institute \\
  Carnegie Mellon University \\
  {\tt \{tianchez, max+\}@cs.cmu.edu}}

\date{}

\begin{document}
\maketitle
\begin{abstract}
  This paper presents an end-to-end framework for task-oriented dialog systems using a variant of Deep Recurrent Q-Networks (DRQN). The model is able to interface with a relational database and jointly learn policies for both language understanding and dialog strategy. Moreover, we propose a hybrid algorithm that combines the strength of reinforcement learning and supervised learning to achieve faster learning speed. We evaluated the proposed model on a 20 Question Game conversational game simulator. Results show that the proposed method outperforms the modular-based baseline and learns a distributed representation of the latent dialog state. 
\end{abstract}

\section{Introduction}
\label{sec:intro}
\begin{figure}[ht!]
    \centering
    \includegraphics[width=0.47\textwidth]{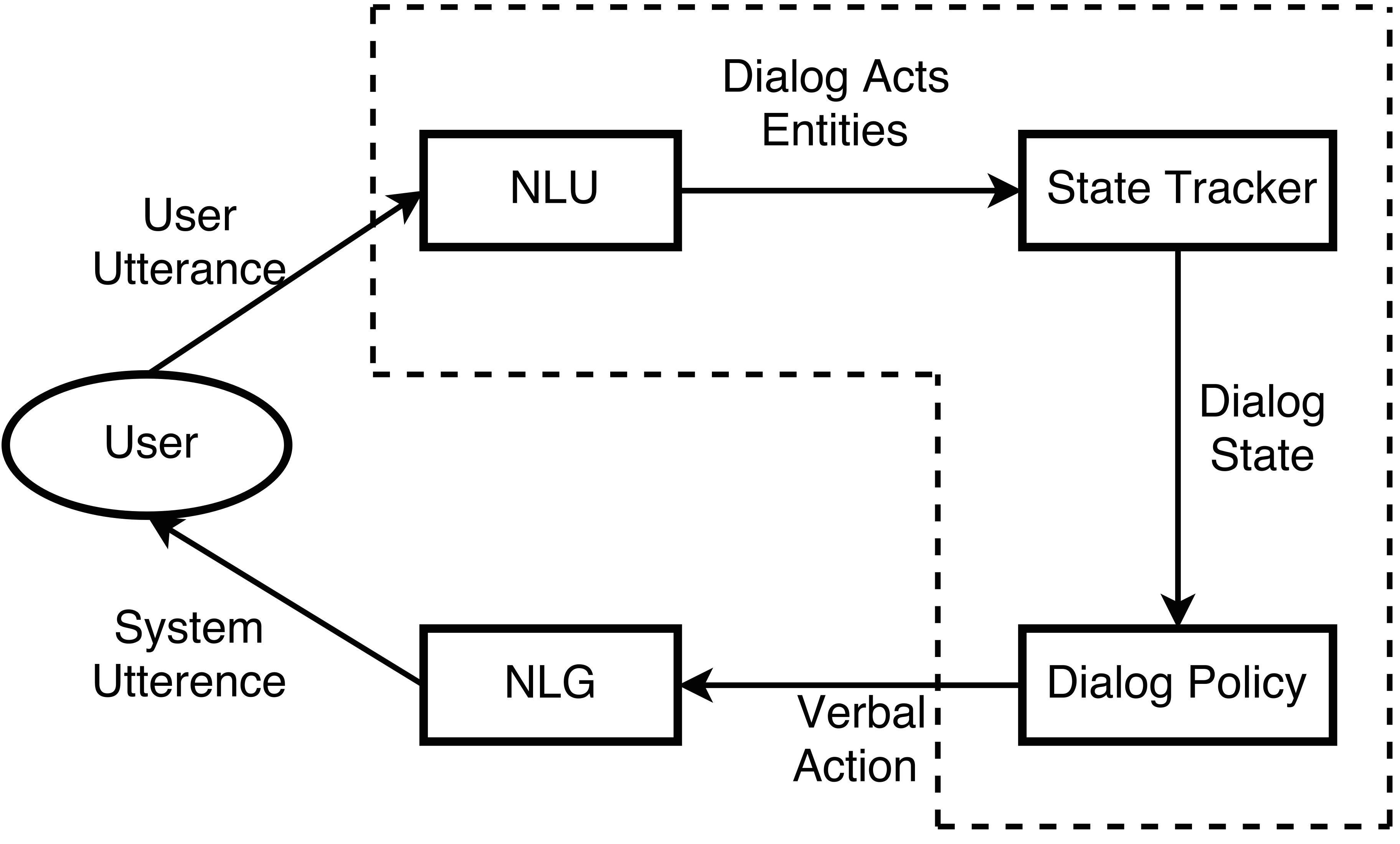}
    \caption{Conventional pipeline of an SDS. The proposed model replaces the modules in the dotted-line box with one end-to-end model.}
    \label{fig:convention}
\end{figure}

Task-oriented dialog systems have been an important branch of spoken dialog system (SDS) research~\cite{raux2005let,young2006using,bohus2003ravenclaw}. The SDS agent has to achieve some predefined targets (e.g. booking a flight) through natural language interaction with the users. The typical structure of a task-oriented dialog system is outlined in Figure~\ref{fig:convention}~\cite{young2006using}. This pipeline consists of several independently-developed modules: natural language understanding (the NLU) maps the user utterances to some semantic representation. This information is further processed by the dialog state tracker (DST), which accumulates the input of the turn along with the dialog history. The DST outputs the current dialog state and the dialog policy selects the next system action based on the dialog state. Then natural language generation (NLG) maps the selected action to its surface form which is sent to the TTS (Text-to-Speech). This process repeats until the agent's goal is satisfied.

The conventional SDS pipeline has limitations. The first issue is the \textit{credit assignment problem}. Developers usually only get feedback from the end users, who inform them about system performance quality. Determining the source of the error requires tedious error analysis in each module because errors from upstream modules can propagate to the rest of the pipeline. The second limitation is \textit{process interdependence}, which makes online adaptation challenging. For example, when one module (e.g. NLU) is retrained with new data, all the others (e.g DM) that depend on it become sub-optimal due to the fact that they were trained on the output distributions of the older version of the module. Although the ideal solution is to retrain the entire pipeline to ensure global optimality, this requires significant human effort. 

Due to these limitations, the goal of this study is to develop an end-to-end framework for task-oriented SDS that replaces 3 important modules: the NLU, the DST and the dialog policy with a single module that can be jointly optimized. Developing such a model for task-oriented dialog systems faces several challenges. The foremost challenge is that a task-oriented system must learn a strategic dialog policy that can achieve the goal of a given task which is beyond the ability of standard supervised learning~\cite{li2014temporal}. The second challenge is that often a task-oriented agent needs to interface with structured external databases, which have symbolic query formats (e.g. SQL query). In order to find answers to the users' requests from the databases, the agent must formulate a valid database query. This is difficult for conventional neural network models which do not provide intermediate symbolic representations.

This paper describes a deep reinforcement learning based end-to-end framework for both dialog state tracking and dialog policy that addresses the above-mentioned issues. We evaluated the proposed approach on a conversational game simulator that requires both language understanding and strategic planning. Our studies yield promising results 1) in jointly learning policies for state tracking and dialog strategies that are superior to a modular-based baseline, 2) in efficiently incorporating various types of labelled data and 3) in learning dialog state representations.

Section~\ref{sec:prior} of the paper discusses related work; Section~\ref{sec:reinforce} reviews the basics of deep reinforcement learning; Section~\ref{sec:model} describes the proposed framework; Section~\ref{sec:analysis} gives experimental results and model analysis; and Section~\ref{sec:conclude} concludes.

\section{Related Work}
\label{sec:prior}
\textbf{Dialog State Tracking:} The process of constantly representing the state of the dialog is called dialog state tracking (DST). Most industrial systems use rule-based heuristics to update the dialog state by selecting a high-confidence output from the NLU~\cite{williams2013dialog}. Numerous advanced statistical methods have been proposed to exploit the correlation between turns to make the system more robust given the uncertainty of the automatic speech recognition (ASR) and the NLU~\cite{bohus2006k,thomson2010bayesian}. The Dialog State Tracking Challenge (DSTC)~\cite{williams2013dialog} formalizes the problem as a supervised sequential labelling task where the state tracker estimates the true slot values based on a sequence of NLU outputs. In practice the output of the state tracker is used by a different dialog policy, so that the distribution in the training data and in the live data are mismatched~\cite{williams2013dialog}. Therefore one of the basic assumptions of DSTC is that the state tracker's performance will translate to better dialog policy performance. Lee~\shortcite{lee2014extrinsic} showed positive results following this assumption by showing a positive correlation between end-to-end dialog performance and state tracking performance.

\textbf{Reinforcement Learning (RL):} RL has been a popular approach for learning the optimal dialog policy of a task-oriented dialog system~\cite{singh2002optimizing,williams2007partially,georgila2011reinforcement,lee2012pomdp}. A dialog policy is formulated as a Partially Observable Markov Decision Process (POMDP) which models the uncertainty existing in both the users' goals and the outputs of the ASR and the NLU. Williams~\shortcite{williams2007partially} showed that POMDP-based systems perform significantly better than rule-based systems especially when the ASR word error rate (WER) is high. Other work has explored methods that improve the amount of training data needed for a POMDP-based dialog manager. Ga{\v{s}}i{\'c}~\shortcite{gavsic2010gaussian} utilized Gaussian Process RL algorithms and greatly reduced the data needed for training. Existing applications of RL to dialog management assume a given dialog state representation. Instead, our approach learns its own dialog state representation from the raw dialogs along with a dialog policy in an end-to-end fashion.

\textbf{End-to-End SDSs:} There have been many attempts to develop end-to-end chat-oriented dialog systems that can directly map from the history of a conversation to the next system response~\cite{vinyals2015neural,serban2015building,shang2015neural}. These methods train sequence-to-sequence models~\cite{sutskever2014sequence} on large human-human conversation corpora. The resulting models are able to do basic chatting with users. The work in this paper differs from them by focusing on building a task-oriented system that can interface with structured databases and provide real information to users. 

Recently, Wen el al.~\shortcite{wen2016network} introduced a network-based end-to-end trainable tasked-oriented dialog system. Their approach treated a dialog system as a mapping problem between the dialog history and the system response. They learned this mapping via a novel variant of the encoder-decoder model. The main differences between our models and theirs are that ours has the advantage of learning a strategic plan using RL and jointly optimizing state tracking beyond standard supervised learning.

\section{Deep Reinforcement Learning}
\label{sec:reinforce}
Before describing the proposed algorithms, we briefly review deep reinforcement learning (RL). RL models are based on the Markov Decision Process (MDP). An MDP is a tuple $(S, A, P, \gamma, R)$, where $S$ is a set of states; $A$ is a set of actions; $P$ defines the transition probability $P(s'|s,a)$; $R$ defines the expected immediate reward $R(s,a)$; and $\gamma \in [0, 1)$ is the discounting factor. The goal of reinforcement learning is to find the optimal policy $\pi^*$, such that the expected cumulative return is maximized~\cite{sutton1998introduction}. MDPs assume full observability of the internal states of the world, which is rarely true for real-world applications. The Partially Observable Markov Decision Process (POMDP) takes the uncertainty in the state variable into account. A POMDP is defined by a tuple $(S, A, P, \gamma, R, O, Z)$. $O$ is a set of observations and $Z$ defines an observation probability $P(o|s,a)$. The other variables are the same as the ones in MDPs. Solving a POMDP usually requires computing the belief state $b(s)$, which is the probability distribution of all possible states, such that $\sum_{s}b(s) = 1$. It has been shown that the belief state is sufficient for optimal control~\cite{monahan1982state}, so that the objective is to find $\pi^*: b \rightarrow a$ that maximizes the expected future return.

\subsection{Deep Q-Network}
The deep Q-Network (DQN) introduced by Mnih~\shortcite{mnih2015human} uses a deep neural network (DNN) to parametrize the Q-value function $Q(s,a;\theta)$ and achieves human-level performance in playing many Atari games. DQN keeps two separate models: a target network $\theta_i^-$ and a behavior network $\theta_i$. For every K new samples, DQN uses $\theta_i^-$ to compute the target values $y^{DQN}$ and updates the parameters in $\theta_i$. Only after every $C$ updates, the new weights of $\theta_i$ are copied over to $\theta_i^-$. Furthermore, DQN utilizes \textit{experience replay} to store all previous experience tuples $(s,a,r,s')$. Before a new model update, the algorithm samples a mini-batch of experiences of size $M$ from the memory and computes the gradient of the following loss function: 
\begin{align}
\mathcal{L}(\theta_i) &=  E_{(s,a,r,s')}[(y^{DQN} - Q(s,a;\theta_i))^2] \\
y^{DQN} &= r + \gamma \max \limits_{a'} Q(s', a' ; \theta_i^-)
\end{align}
Recently, Hasselt et al.~\shortcite{van2015deep} leveraged the overestimation problem of standard Q-Learning by introducing double DQN and Schaul et al.~\shortcite{schaul2015prioritized} improves the convergence speed of DQN via \textit{prioritized experience replay}. We found both modifications useful and included them in our studies.

\subsection{Deep Recurrent Q-Network}
An extension to DQN is a Deep Recurrent Q-Network (DRQN) which introduces a Long Short-Term Memory (LSTM) layer~\cite{hochreiter1997long}  on top of the convolutional layer of the original DQN model~\cite{hausknecht2015deep} which allows DRQN to solve POMDPs. The recurrent neural network can thus be viewed as an approximation of the belief state that can aggregate information from a sequence of observations. Hausknecht~\shortcite{hausknecht2015deep} shows that DRQN performs significantly better than DQN when an agent only observes partial states. A similar model was proposed by Narasimhan and Kulkarni~\shortcite{narasimhan2015language} and learns to play Multi-User Dungeon (MUD) games~\cite{curtis1992mudding} with game states hidden in natural language paragraphs.

\section{Proposed Model}
\label{sec:model}
\subsection{Overview}
End-to-end learning refers to models that can back-propagate error signals from the end output to the raw inputs. Prior work in end-to-end state tracking~\cite{henderson2014word} learns a sequential classifier that estimates the dialog state based on ASR output without the need of an NLU. Instead of treating state tracking as a standard supervised learning task, we propose to unify dialog state tracking with the dialog policy so that both are treated as actions available to a reinforcement learning agent. Specifically, we learn an optimal policy that either generates a verbal response or modifies the current estimated dialog state based on the new observations. This formulation makes it possible to obtain a state tracker even without the labelled data required for DSTC, as long as the rewards from the users and the databases are available. Furthermore, in cases where dialog state tracking labels are available, the proposed model can incorporate them with minimum modification and greatly accelerate its learning speed. Thus, the following sections describe two models: RL and Hybrid-RL, corresponding to two labelling scenarios: 1) only dialog success labels and 2) dialog success labels with state tracking labels.

\begin{figure}
    \centering
    \includegraphics[width=0.47\textwidth]{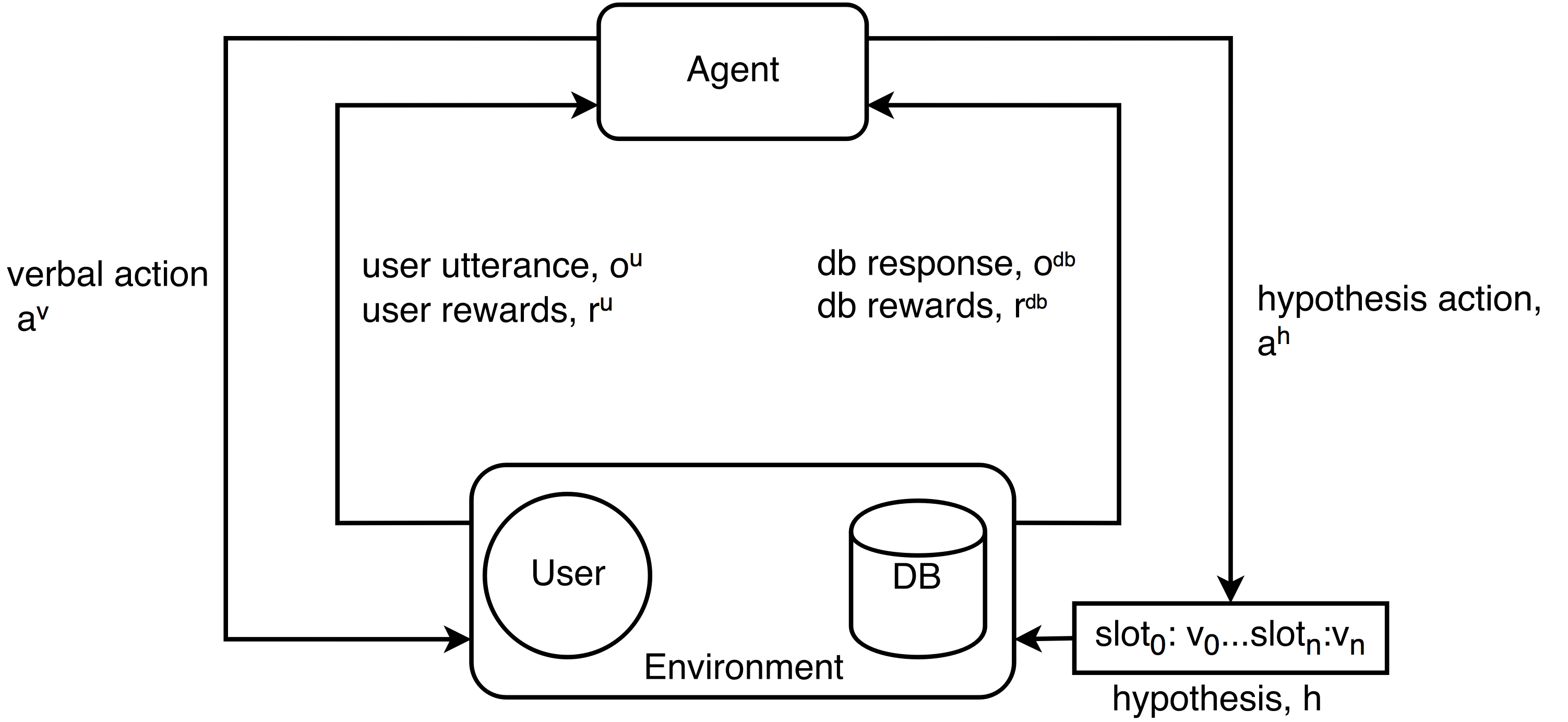}
    \caption{An overview of the proposed end-to-end task-oriented dialog management framework.}
    \label{fig:abstract}
\end{figure}

\subsection{Learning from the Users and Databases}
Figure~\ref{fig:abstract} shows an overview of the framework. We consider a task-oriented dialog task, in which there are $S$ slots, each with cardinality $C_i, i \in [0, S)$. The environment consists of a user, $E^u$ and a database $E^{db}$. The agent can send verbal actions, $a^v \in A^v$ to the user and the user will reply with natural language responses $o^u$ and rewards $r^u$. In order to interface with the database environment $E^{db}$, the agent can apply special actions $a^h \in A^h$ that can modify a query hypothesis $h$. The hypothesis is a slot-filling form that represents the most likely slot values given the observed evidence. Given this hypothesis, $h$, the database can perform a normal query and give the results as observations, $o^{db}$ and rewards $r^{db}$. 

\begin{figure}
    \centering
    \includegraphics[width=0.47\textwidth]{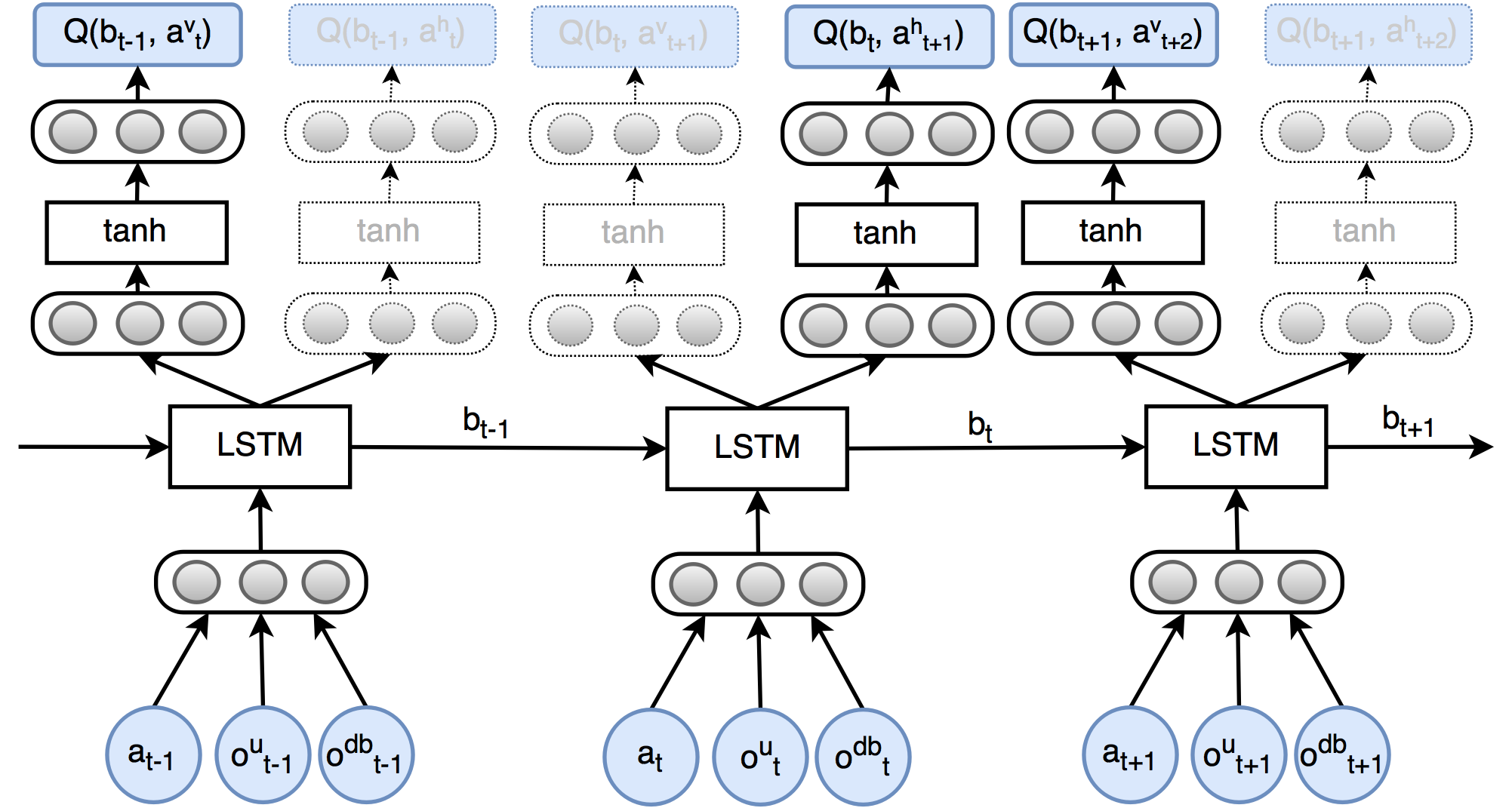}
    \caption{The network takes the observation $o_t$ at turn $t$. The recurrent unit updates its hidden state based on both the history and the current turn embedding. Then the model outputs the Q-values for all actions. The policy network in grey is masked by the action mask}
    \label{fig:network}
\end{figure}

At each turn $t$, the agent applies its selected action $a_t \in \{A^v, A^h\}$ and receives the observations from either the user or the database. We can then define the observation $o^t$ of turn $t$ as,
\begin{align}
o^t &= \begin{bmatrix}
           a_t \\
           o_t^u \\
           o_t^{db}
         \end{bmatrix}
\end{align}
We then use the LSTM network as the dialog state tracker that is capable of aggregating information over turns and generating a dialog state representation, $b_t = LSTM (o_t, b_{t-1})$, where $b_t$ is an approximation of the belief state at turn $t$. Finally, the dialog state representation from the LSTM network is the input to $S+1$ policy networks implemented as Multilayer Perceptrons (MLP). The first policy network approximates the Q-value function for all verbal actions $Q(b_t, a^v)$ while the rest estimate the Q-value function for each slot, $Q(b_t, a^h)$, as shown in Figure~\ref{fig:network}.

\subsection{Incorporating State Tracking Labels}
The pure RL approach described in the previous section could suffer from slow convergence when the cardinality of slots is large. This is due to the nature of reinforcement learning: that it has to try different actions (possible values of a slot) in order to estimate the expected long-term payoff. On the other hand, a supervised classifier can learn much more efficiently. A typical multi-class classification loss function (e.g. categorical cross entropy) assumes that there is a single correct label such that it encourages the probability of the correct label and suppresses the probabilities of the all the wrong ones. Modeling dialog state tracking as a Q-value function has advantages over a local classifier. For instance, take the situation where a user wants to send an email and the state tracker needs to estimate the user's goal from among three possible values: \textit{send}, \textit{edit} and \textit{delete}. In a classification task, all the incorrect labels (\textit{edit}, \textit{delete}) are treated as equally undesirable. However, the cost of mistakenly recognizing the user goal as \textit{delete} is much larger than \textit{edit}, which can only be learned from the future rewards. In order to train the slot-filling policy with both short-term and long-term supervision signals, we decompose the reward function for $A^h$ into two parts:
\begin{align}
    Q^{\pi}(b, a^h) &= \bar{R}(b,a) + \gamma \sum_{b'}P(b'|b,a^h) V^{\pi}(b') \\ 
     \bar{R}(b,a,b') &= R(b, a^h) + P(a^h|b)
\end{align}
where $P(a^h|b)$ is the conditional probability that the correct label of the slots is $a^h$ given the current belief state. In practice, instead of training a separate model estimating $P(a^h|b)$, we can replace $P(a^h|b)$ by $\mathds{1}(y=a^h)$ as the sample reward $r$, where $y$ is the label. Furthermore, a key observation is that although it is expensive to collect data from the user $E^u$,  one can easily sample trajectories of interaction with the database since $P(b'|b,a^h)$ is known. Therefore, we can accelerate learning by generating synthetic experiences, i.e. tuple $(b, a^h, r, b') \forall a^h \in A^h$ and add them to the experience replay buffer. This approach is closely related to the Dyna Q-Learning proposed in~\cite{sutton1990integrated}. The difference is that Dyna Q-learning uses the estimated environment dynamics to generating experiences, while our method only uses the known transition function (i.e. the dynamics of the database) to generate synthetic samples.

\subsection{Implementation Details}
We can optimize the network architecture in several ways to improve its efficiency:

\textbf{Shared State Tracking Policies:} it is more efficient to tie the weights of the policy networks for similar slots  and use the index of slot as an input. This can reduce the number of parameters that needs to be learned and encourage shared structures. The studies in Section~\ref{sec:analysis} illustrate one example. 

\textbf{Constrained Action Mask:} We can constrain the available actions at each turn to force the agent to alternate between verbal response and slot-filling. We define $A_{mask}$ as a function that takes state $s$ and outputs a set of available actions for:
\begin{align}
    A_{mask}(s) &= A_h \quad \text{new inputs from the user} \\
                &= A_v \quad \text{otherwise}
\end{align}

\textbf{Reward Shaping based on the Database:} the reward signals from the users are usually sparse (at the end of a dialog), the database, however, can provide frequent rewards to the agent. Reward shaping is a technique used to speed up learning. Ng et al.~\shortcite{ng1999policy} showed that potential-based reward shaping does not alter the optimal solution; it only impacts the learning speed. The pseudo reward function $F(s,a,s')$ is defined as:
\begin{align}
\bar{R}(s,a,s') &= R(s,a,s') + F(s,a,s') \\
F(s,a,s') &= \gamma \phi(s') - \phi(s)
\end{align}

Let the total number of entities in the database be $D$ and $P_{max}$ be the max potential, the potential $\phi(s)$ is:
\begin{align}
    \phi(s_t) &= P_{max} (1- \frac{d_t}{D}) \quad \text{if $d_t > 0$} \\
    \phi(s_t) &= 0 \quad \text{if $d_t = 0$}
\end{align}

The intuition of this potential function is to encourage the agent to narrow down the possible range of valid entities as quickly as possible. Meanwhile, if no entities are consistent with the current hypothesis, this implies that there are mistakes in previous slot filling, which gives a potential of 0.

\section{Experiments}
\label{sec:analysis}
\subsection{20Q Game as Task-oriented Dialog}
In order to test the proposed framework, we chose the 20 Question Game (20Q). The game rules are as follows: at the beginning of each game, the user thinks of a famous person. Then the agent asks the user a series of Yes/No questions. The user honestly answers, using one of three answers: yes, no or I don't know. In order to have this resemble a dialog, our user can answer with any natural utterance representing one of the three intents. The agent can make guesses at any turn, but a wrong guess results in a negative reward. The goal is to guess the correct person within a maximum number of turns with the least number of wrong guesses. An example game conversation is as follows:
\begin{addmargin}[1em]{2em}
\textit{\textbf{Sys}: Is this person male?\\
\textbf{User}: Yes I think so.\\
\textbf{Sys}: Is this person an artist?\\
\textbf{User}: He is not an artist.\\
... \\
\textbf{Sys}: I guess this person is Bill Gates.\\
\textbf{User}: Correct.
}
\end{addmargin}

We can formulate the game as a slot-filling dialog. Assume the system has $|Q|$ available questions to select from at each turn. The answer to each question becomes a slot and each slot has three possible values: \textit{yes/no/unknown}. Due to the length limit and wrong guess penalty, the optimal policy does not allow the agent to ask all of the questions regardless of the context or guess every person in the database one by one.

\subsection{Simulator Construction}
We constructed a simulator for 20Q. The simulator has two parts: a database of 100 famous people and a user simulator. 

We selected 100 people from Freebase~\cite{bollacker2008freebase}, each of them has 6 attributes: \textit{birthplace, degree, gender, profession and birthday}. We manually designed several Yes/No questions for each attribute that is available to the agent. Each question covers a different set of possible values for a given attribute and thus carries a different discriminative power to pinpoint the person that the user is thinking of. As a result, the agent needs to judiciously select the question, given the context of the game, in order to narrow down the range of valid people. There are 31 questions. Table~\ref{tbl:questions} shows a summary.
\begin{table}[ht!]
    \centering
    \begin{tabular}{p{0.1\textwidth}p{0.015\textwidth}p{0.29\textwidth}} \hline
    Attribute & $Q_a$ & Example Question \\ \hline
    Birthday   & 3 & Was he/she born before 1950? \\
    Birthplace & 9 & Was he/she born in USA? \\ 
    Degree     & 4 & Does he/she have a PhD? \\ 
    Gender     & 2 & Is this person male? \\ 
    Profession & 8 & Is he/she an artist? \\ 
    Nationality & 5 & Is he/she a citizen of an Asian country? \\ \hline
    \end{tabular}
    \caption{Summary of the available questions. $Q_a$ is the number of questions for attribute $a$.\\}
    \label{tbl:questions}
\end{table}

At the beginning of each game, the simulator will first uniformly sample a person from the database as the person it is thinking of. Also there is a $5\%$ chance that the simulator will consider \textit{unknown} as an attribute and thus it will answer with \textit{unknown} intent for any question related to it. After the game begins, when the agent asks a question, the simulator first determines the answer (\textit{yes, no or unknown}) and replies using natural language. In order to generate realistic natural language with the \textit{yes/no/unknown} intent, we collected utterances from the Switchboard Dialog Act (SWDA) Corpus~\cite{jurafsky1997switchboard}. Table~\ref{tbl:natural_resp} presents the mapping from the SWDA dialog acts to \textit{yes/no/unknown}. We further post-processed results and removed irrelevant utterances, which led to 508, 445 and 251 unique utterances with intent respectively \textit{yes/no/unknown}. We keep the frequency counts for each unique expression. Thus at run time, the simulator can sample a response according to the original distribution in the SWDA Corpus.

\begin{table}[ht!]
    \centering
    \begin{tabular}{p{0.1\textwidth}p{0.3\textwidth}} \hline
    Intent & SWDA tags \\ \hline
    Yes   & Agree, Yes answers, Affirmative non-yes answers \\\hline
    No    & No answers, Reject, Negative non-no answers\\ \hline
    Unknown & Maybe, Other Answer\\ \hline
    \end{tabular}
    \caption{Dialog act mapping from SWDA to \textit{yes/no/unknown}}
    \label{tbl:natural_resp}
\end{table}

A game is terminated when one of the four conditions is fulfilled: 1) the agent guesses the correct answer, 2) there are no people in the database consistent with the current hypothesis, 3) the max game length (100 steps) is reached and 4) the max number of guesses is reached (10 guesses). Only if the agent guesses the correct answer (condition 1) treated as a game victory. The win and loss rewards are $30$ and $-30$ and a wrong guess leads to a $-5$ penalty.
\subsection{Training Details}
The user environment $E^{u}$ is the simulator that only accepts verbal actions, either a Yes/No question or a guess, and replies with a natural language utterance. Therefore $A^v$ contains $|Q|+1$ actions, in which the first $|Q|$ actions are questions and the last action makes a guess, given the results from the database.

The database environment reads in a query hypothesis $h$ and returns a list of people that satisfy the constraints in the query. $h$ has a size of $|Q|$ and each dimension can be one of the three values: \textit{yes/no/unknown}. Since the cardinality for all slots is the same, we only need 1 slot-filling policy network with 3 Q-value outputs for \textit{yes/no/unknown}, to modify the value of the latest asked question, which is the shared policy approach mentioned in Section~\ref{sec:model}. Thus $A^h = \{yes, no,unknown\}$. For example, considering $Q=3$ and the hypothesis $h$ is: $[unknown, unknown, unknown]$. If the latest asked question is $Q_1$ (1-based), then applying action $a^h=yes$ will result in the new hypothesis: $[yes, unknown, unknown]$. 

To represent the observation $o_t$ in vectorial form, we use a bag-of-bigrams feature vector to represent a user utterance; a one-hot vector to represent a system action and a single discrete number to represent the number of people satisfying the current hypothesis. 

The hyper-parameters of the neural network model are as follows: the size of turn embedding is 30; the size of LSTMs is 256; each policy network has a hidden layer of 128 with $tanh$ activation. We also add a dropout rate of 0.3 for both LSTMs and $tanh$ layer outputs. The network has a total of 470,005 parameters. The network was trained through $RMSProp$~\cite{tieleman2012lecture}. For hyper-parameters of DRQN, the behavior network was updated every 4 steps and the interval between each target network update $C$ is 1000. $\epsilon$-greedy exploration is used for training, where $\epsilon$ is linearly decreased from 1 to 0.1. The reward shaping constant $P_{max}$ is 2 and the discounting factor $\gamma$ is $0.99$. The resulting network was evaluated every 5000 steps and the model was trained up to 120,000 steps. Each evaluation records the agent's performance with a greedy policy for 200 independent episodes.
\subsection{Dialog Policy Analysis}
We compare the performance of three models: a strong modular baseline, RL and Hybrid-RL. The baseline has an independently trained state tracker and dialog policy. The state tracker is also an LSTM-based classifier that inputs a dialog history and predicts the slot-value of the latest question. The dialog policy is a DRQN that assumes perfect slot-filling during training and simply controls the next verbal action. Thus the essential difference between the baseline and the proposed models is that the state tracker and dialog policy are not trained jointly. Also, since hybrid-RL effectively changes the reward function, the typical average cumulative reward metric is not applicable for performance comparison. Therefore, we directly compare the win rate and average game length in later discussions. 
\begin{table}[ht!]
    \centering
    \begin{tabular}{p{0.11\textwidth}|p{0.14\textwidth}p{0.14\textwidth}} \hline 
             & Win Rate (\%) & Avg Turn \\ \hline
    Baseline & 68.5 & 12.2 \\
    RL       & 85.6 & 21.6 \\ 
    Hybrid-RL & 90.5 & 19.22\\ \hline
    \end{tabular}
    \caption{Performance of the three systems}
    \label{tbl:best}
\end{table}

\begin{figure}[ht!]
    \centering
    \includegraphics[width=0.47\textwidth]{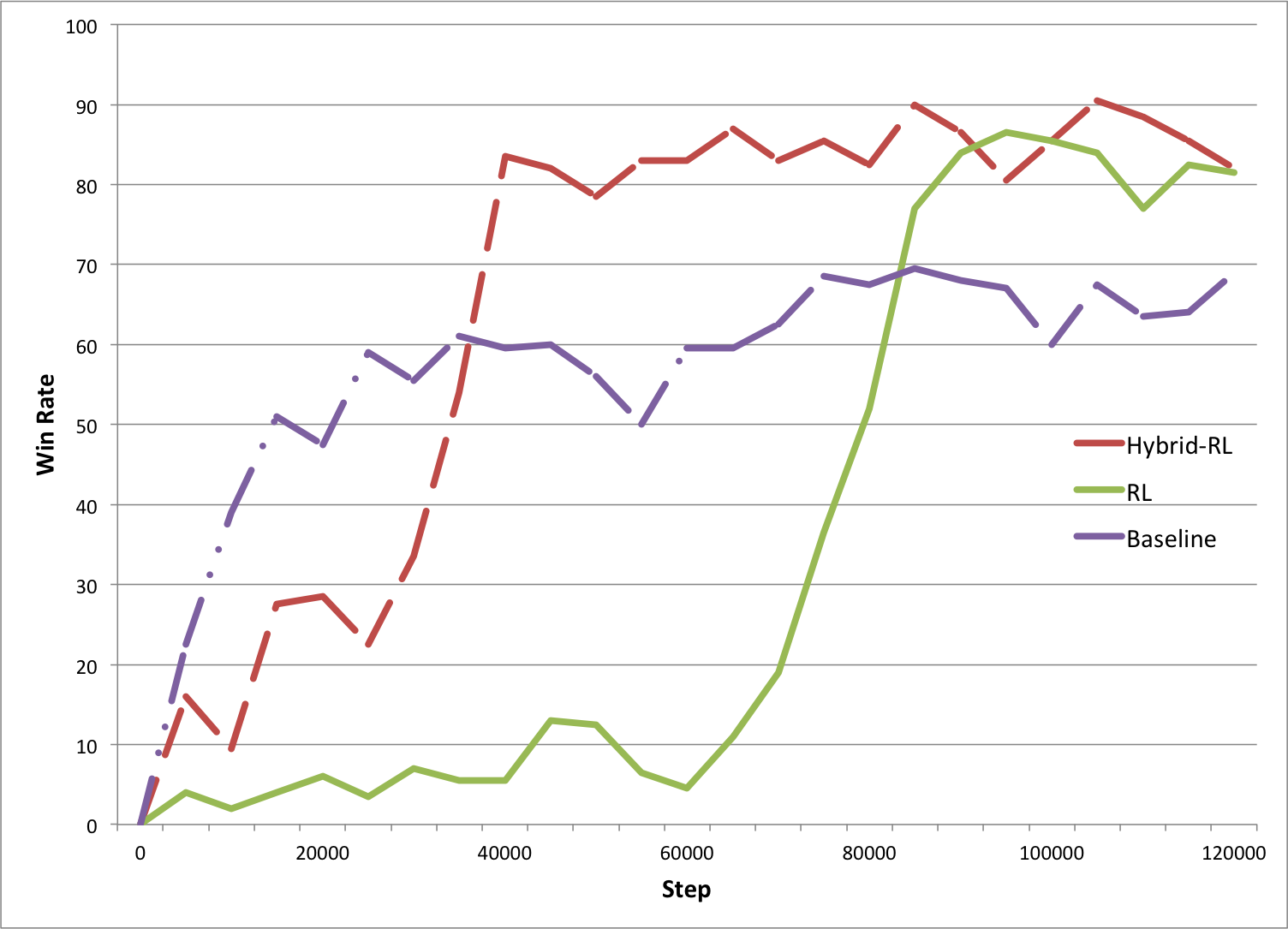}
    \caption{Graphs showing the evolution of the win rate during training.}
    \label{fig:winrate}
\end{figure}

Table~\ref{tbl:best} shows that both proposed models achieve significantly higher win rate than the baseline by asking more questions before making guesses. Figure~\ref{fig:winrate} illustrates the learning process of the three models. The horizontal axis is the total number of interaction between the agent and either the user or the database. The baseline model has the fastest learning speed but its performance saturated quickly because the dialog policy was not trained together with the state tracker. So the dialog policy is not aware of the uncertainty in slot-filling and the slot-filler does not distinguish between the consequences of different wrong labels (e.g classify \textit{yes} to \textit{no} versus to \textit{unknown}). On the other hand, although RL reaches high performance at the end of the training, it struggles in the early stages and suffers from slow convergence. This is due to that fact that correct slot-filling is a prerequisite for winning 20Q, while the reward signal has a long delayed horizon in the RL approach. Finally, the hybrid-RL approach is able to converge to the optimal solution much faster than RL due to the fact that it efficiently exploits the information in the state tracking label.

\subsection{State Tracking Analysis}
One of the hypotheses is that the RL approach can learn a good state tracker using only dialog success reward signals. We ran the best trained models using a greedy policy and collected 10,000 samples. Table~\ref{tbl:metric} reports the precision and recall of slot filling in these trajectories.
\begin{table}[ht!]
    \centering
    \begin{tabular}{p{0.11\textwidth}|p{0.1\textwidth}p{0.09\textwidth}p{0.09\textwidth}} \hline 
              & Unknown & Yes & No \\ \hline
    Baseline & 0.99/0.60 & 0.96/0.97 & 0.94/0.95 \\
    RL & 0.21/0.77 & 1.00/0.93 & 0.95/0.51 \\
    Hybrid-RL & 0.54/0.60 & 0.98/0.92 & 0.94/0.93 \\ \hline
    \end{tabular}
    \caption{State tracking performance of the three systems. The results are in the format of \textit{precision/recall}}
    \label{tbl:metric}
\end{table}
The results indicate that the RL model learns a completely different strategy compared to the baseline. The RL model aims for high precision so that it predicts \textit{unknown} when the input is ambiguous, which is a safer option than predicting \textit{yes/no}, because confusing between \textit{yes} and \textit{no} may potentially lead to a contradiction and a game failure. This is very different from the baseline which does not distinguish between incorrect labels. Therefore, although the baseline achieves better classification metrics, it does not take into account the long-term payoff and performs sub-optimally in terms of overall performance.

\subsection{Dialog State Representation Analysis}
Tracking the state over multiple turns is crucial because the agent's optimal action depends on the history, e.g. the question it has already asked, the number of guesses it has spent. Furthermore, one of the assumptions is that the output of the LSTM network is an approximation of the belief state in the POMDP. We conducted two studies to test these hypotheses. For both studies, we ran the Hybrid-RL models saved at 20K, 50K and 100K steps against the simulator with a greedy policy and recorded 10,000 samples for each model. 

The first study checks whether we can reconstruct an important state feature: the number of guesses the agent has made from the dialog state embedding. We divide the collected 10,000 samples into 80\% for training and 20\% for testing. We used the LSTM output as input features to a linear regression model with $l2$ regularization. Table~\ref{tbl:state_regression} shows the correlation of determination $r^2$ increases for the model that was trained with more data. 

\begin{table}[ht!]
    \centering
    \begin{tabular}{p{0.1\textwidth}|p{0.09\textwidth}p{0.09\textwidth}p{0.09\textwidth}} \hline
    Model & 20K & 50K & 100K \\ \hline
    $r^2$ & 0.05 & 0.51 & 0.77 \\  \hline
    \end{tabular}
    \caption{$r^2$ of the linear regression for predicting the number of guesses in the test dataset.}
    \label{tbl:state_regression}
\end{table}

The second study is a retrieval task. The latent state of the 20Q game is the true intent of the users' answers to all the questions that have been asked so far. Therefore, the true state vector, $s$ has a size of 31 and each slot, $s[k], k \in [0, 31)$ is one of the four values: \textit{not yet asked, yes, no, unknown}.  Therefore, if the LSTM output $b$ is in fact implicitly learning the distribution over this latent state $s$, they must be highly correlated for a well-trained model. Therefore, for each $b_i, i \in [0, 10,000)$, we find its nearest 5 neighbors based on cosine distance measuring and record their latent states, $N(b_i): B \rightarrow [S] $. Then we compute the empirical probability that each slot of the true state $s$ differs from the retrieved neighbors:
\begin{equation}
    p_{\text{diff}}(s[k]) = E_{s_i}\bigg[\frac{\sum_{n=0}^{4} \mathds{1}(N(b_i)[n][k] \neq s_i[k])}{5}\bigg]\\
\end{equation}
where $\mathds{1}$ is the indicator function, $k$ is the slot index and $n$ is the neighbor index.

\begin{figure}[ht!]
    \centering
    \includegraphics[width=0.47\textwidth]{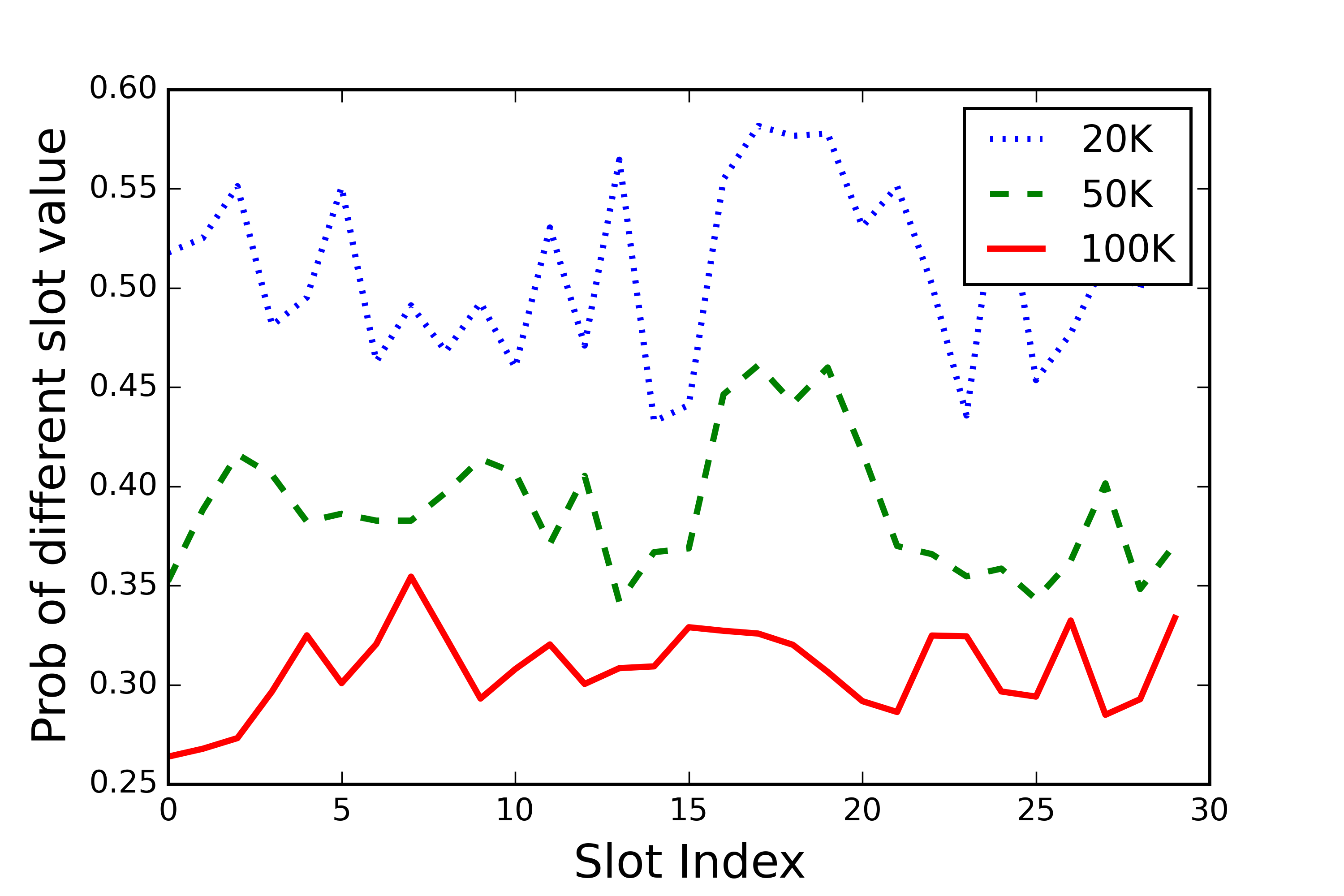}
    \caption{Performance of retrieving similar true dialog states using learned dialog state embedding.}
    \label{fig:retrival}
\end{figure}

Figure~\ref{fig:retrival} shows that the retrieval error continues to decrease for the model that was trained better, which confirms our assumption that the LSTM output is an approximation of the belief state.

\section{Conclusion}
\label{sec:conclude}
This paper identifies the limitations of the conventional SDS pipeline and describes a novel end-to-end framework for a task-oriented dialog system using deep reinforcement learning. We have assessed the model on the 20Q game. The proposed models show superior performance for both natural language understanding and dialog strategy. Furthermore, our analysis confirms our hypotheses that the proposed models implicitly capture essential information in the latent dialog states.

One limitation of the proposed approach is poor scalability due to the large number of samples needed for convergence. So future studies will include developing full-fledged task-orientated dialog systems and exploring methods to improve the sample efficiency. Also, investigating techniques that allow easy integration of domain knowledge so that the system can be more easily debugged and corrected is another important direction.

\section{Acknowledgements}
This work was funded by NSF grant CNS-1512973. The opinions expressed in this paper do not necessarily reflect those of NSF. We would also like to thank Alan W Black for discussions on this paper.

\bibliographystyle{acl2016}
\bibliography{acl2016}

\end{document}